\pdfoutput=1

\documentclass[11pt]{article}

\usepackage[]{acl}

\usepackage{times}
\usepackage{latexsym}

\usepackage[T1]{fontenc}

\usepackage[utf8]{inputenc}

\usepackage{microtype}

\usepackage{inconsolata}

\usepackage{url}
\usepackage{array}
\usepackage{xspace}
\usepackage{xcolor}
\usepackage{lipsum}
\usepackage{wrapfig}
\usepackage{caption}
\usepackage{hyperref}
\usepackage{enumitem}
\usepackage{booktabs}
\usepackage{multirow}
\usepackage{graphicx}
\usepackage{listings}
\usepackage{subcaption}
\usepackage{tablefootnote}
\usepackage[most]{tcolorbox}
\usepackage{tabularx}

\usepackage{float}

\usepackage{pgfplots}
\usepackage{xspace}

\newcommand{\todo}[1]{\textcolor{blue}{\textbf{[TODO: #1]}}}
\newcommand{\hide}[1]{}
 \usepackage{graphicx}

\newcommand{\fork}{\texttt{[Fork]}\xspace}
\newcommand{\child}{\texttt{[Child]}\xspace}
\newcommand{\vapar}{Vanilla-APAR\xspace}
\newcommand{\mapar}{Medusa-APAR\xspace}
\newcommand{\bapar}{Batched-APAR\xspace}
\usepackage{algorithm}
\usepackage[noend]{algpseudocode}
\usepgfplotslibrary{groupplots}

\newcommand{\specialfootnote}[1]{\begingroup
    \renewcommand\thefootnote{}\footnote{#1}
    \addtocounter{footnote}{-1}
    \endgroup
}

\definecolor{darkred}{rgb}{0.55, 0.0, 0.0}

\title{APAR: LLMs Can Do Auto-Parallel Auto-Regressive Decoding}

\author{Mingdao Liu$^{1,\dagger,*}$, Aohan Zeng$^{1,2,*}$, Bowen Wang$^{1,\dagger}$, Peng Zhang$^{2}$,  Jie Tang$^{1}$, Yuxiao Dong$^{1}$\\ \\
$^1$Tsinghua University $^2$Zhipu AI}

\begin{document}

\maketitle

\begin{abstract}
The massive adoption of large language models (LLMs) demands efficient deployment strategies. 
However, the auto-regressive decoding process, which is fundamental to how most LLMs generate text, poses challenges to achieve efficient serving. 
In this work, we introduce a parallel auto-regressive generation method. 
By instruct-tuning on general domain data that contains hierarchical structures, we enable LLMs to independently plan their generation process and perform auto-parallel auto-regressive (APAR) generation, significantly reducing the number of generation steps. 
APAR alone can achieve up to 2$\times$ speed-up, and when combined with speculative decoding, the speed-up can reach up to 4$\times$. 
In addition, APAR reduces the key-value cache consumption and attention computation during generation. 
This leads to a throughput increase of 20-70\% and a latency reduce of 20-35\% in high-throughput scenarios, compared to state-of-the-art serving frameworks. 
\end{abstract}

\specialfootnote{$^*$ Equal contribution.}
\specialfootnote{$^\dagger$ Work done while these authors interned at Zhipu AI.}

\section{Introduction}
\label{intro}

Large language models (LLMs) \citep{openai2023gpt4,touvron2023llama,zeng2022glm} have increasingly become foundational to various AI applications~\citep{autogpt,babyagi,Park2023GenerativeAgents,zhou2023webarena}. 
This widespread adoption has led to a growing demand for efficient model deployment, i.e., low latency and high throughput~\citep{deepspeed-inference}. 
However, the intrinsic auto-regressive (AR) structure of these models presents significant challenges in achieving more efficient serving~\citep{radford2018improving}.

\begin{figure}[htb]
\begin{center}
\includegraphics[width=\linewidth]{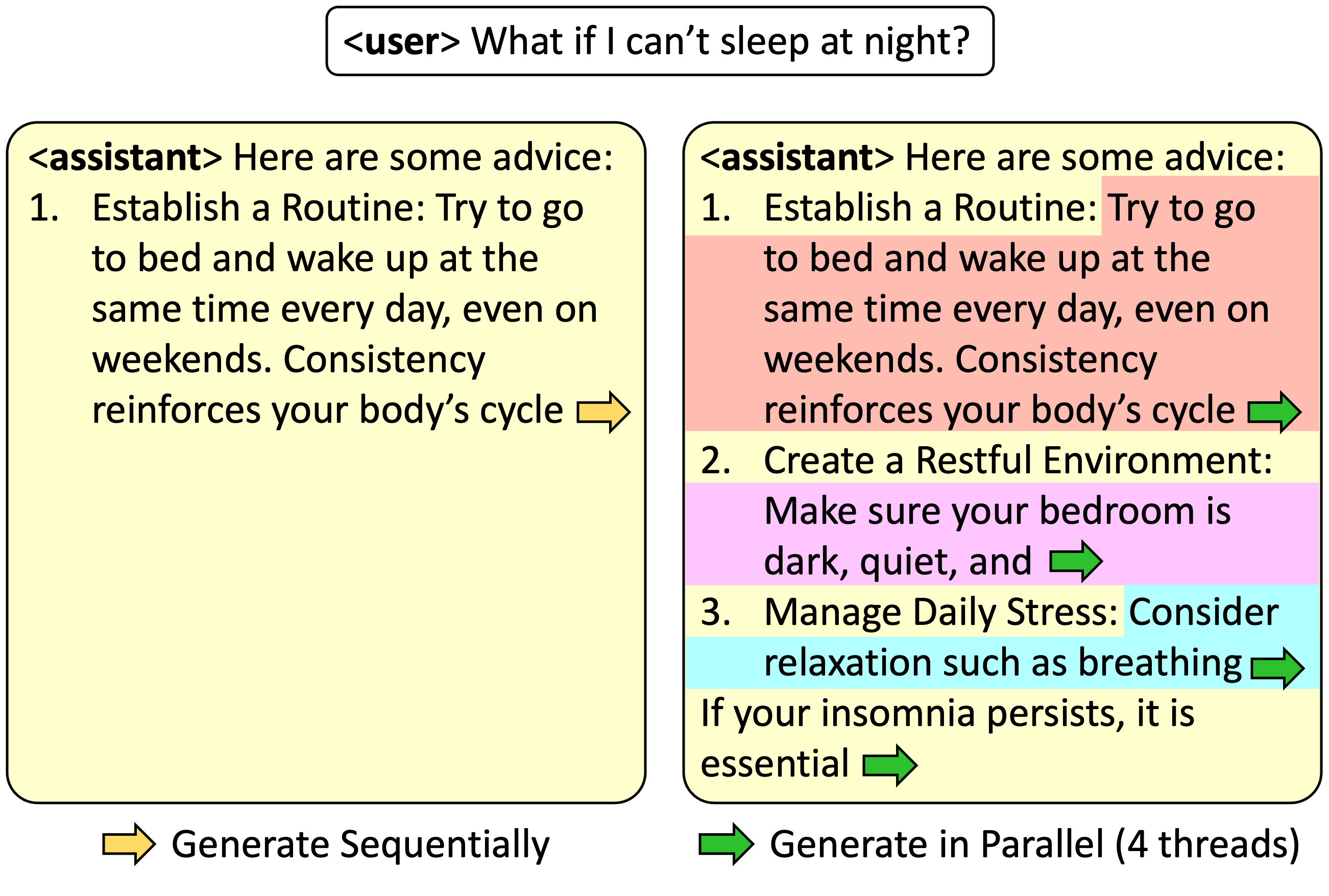} 
\end{center}
\vspace{-0.5cm}
\caption{\textbf{APAR Decoding Overview.} Different from the original auto-regressive decoding, APAR detects potential parts to be generated in parallel and issues multiple generation threads.}
\label{fig:demo}
\end{figure}

First, each new token is auto-regressively generated conditioned on the entire set of previously-generated tokens. 
This incremental decoding process results in sub-optimal generation speeds, as each generation step requires accessing the vast number of parameters of a LLM~\citep{deepspeed-inference}. 
Consequently, when the generation batch size is not sufficiently large, this process becomes memory-bound, resulting in an under-utilization of the GPU compute. 

Second, the computation of attention over all preceding tokens in Transformer \citep{vaswani2017attention} also limits the serving throughput. 
In high-throughput scenarios, many sequences are generating in parallel and the generation process becomes computation-bound.
Meanwhile, the computation cost of attention scales linearly with the sequence length, which hinders further improvements of the throughput, especially for long responses. 
In addition, the caching of key and value tensors (KV cache) for generated tokens, despite advancements in memory-efficient algorithms~\citep{vllm}, scales linearly with the sequence length, constraining the number of concurrent requests that a system can handle.

In light of these challenges, we introduce the Auto-Parallel Auto-Regressive (APAR) decoding strategy with the goal of improving the inference efficiency of LLMs. 
APAR leverages the inherent parallelizable structure 
in LLM generation, capitalizing on LLMs' understanding of text structures. 
By fine-tuning LLMs on corpora with hierarchical structures, the models can learn to autonomously initiate parallel generation threads when encountering parallelizable response structures. 
This approach transforms the conventional linear generation into a parallelizable paragraph tree structure. 
This not only facilitates greater decoding parallelism but also reduces attention spans through tree-based attention mechanisms, and enables the earlier release of consumed KV cache memory. 

We perform experiments on the Vicuna family of models. 
In memory-bound scenarios, APAR can help reduce the model latency and achieve an average generation speed increase of 2$\times$ on Vicuna Bench \citep{vicuna2023}. 
Furthermore, the design of APAR is complementary to most existing inference acceleration methods. 
For example, when combined with Medusa \citep{medusa}, a speculative decoding strategy, APAR-based models yield speed improvements of up to 4$\times$ on Vicuna Bench. 
In certain specific categories, this combination even achieves a speed-up of up to 6$\times$. 

In high-throughput scenarios, APAR's compatibility with vLLM enables early memory release, reducing KV cache requirements by up to 50\% while still maintaining the same level of throughput. 
In addition, APAR reduces the number of tokens involved in attention calculation. 
By using the same amount of KV cache memory, it gets a 20-70\% improvement in throughput over the original AR process, and achieves a 20-35\% reduction in latency while maintaining the same serving concurrency.

Importantly, the quality of generation with APAR is not compromised. 
Evaluations across multiple categories on the MT Bench and Vicuna Bench \citep{mt-bench} demonstrate that the response quality remains largely consistent, with variations within a $\pm$2\% range compared to its AR counterparts. 
This indicates that APAR-based models retain the contextual generation capability while enhancing the decoding speed and efficiency.

\section{Auto-Parallel Auto-Regressive Decoding}
\subsection{Overview}
Parallelizable structures are ubiquitous in the response of LLMs. 
For instance, in the ShareGPT dataset, 58\% of the dialogues contain at least one response of ordered or unordered lists from ChatGPT, and about 32\% of the responses contains listed structure.
Most listed structures are naturally suitable for paralleled generation, since the details of an itemized paragraph is usually conditioned on its leading sentence or phrase. 

\begin{figure*}[t]
\begin{center}
\includegraphics[width=\linewidth]{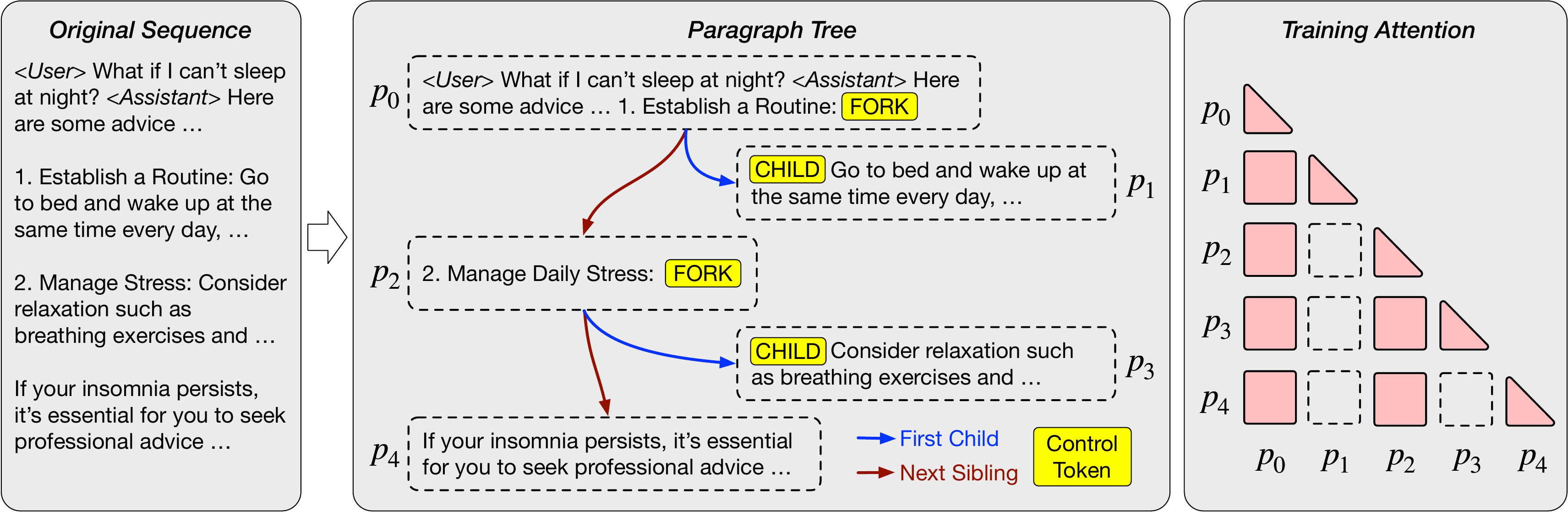}
\end{center}
\caption{\textbf{APAR Training Inputs.} Based on pre-defined rules, the \emph{original sequence} is transformed into a \emph{paragraph tree}, which is used to train APAR models. Any token attends only to tokens on its path to root.}
\label{fig:paragraph_tree}
\end{figure*}

\begin{figure*}[t]
\begin{center}
\includegraphics[width=\linewidth]{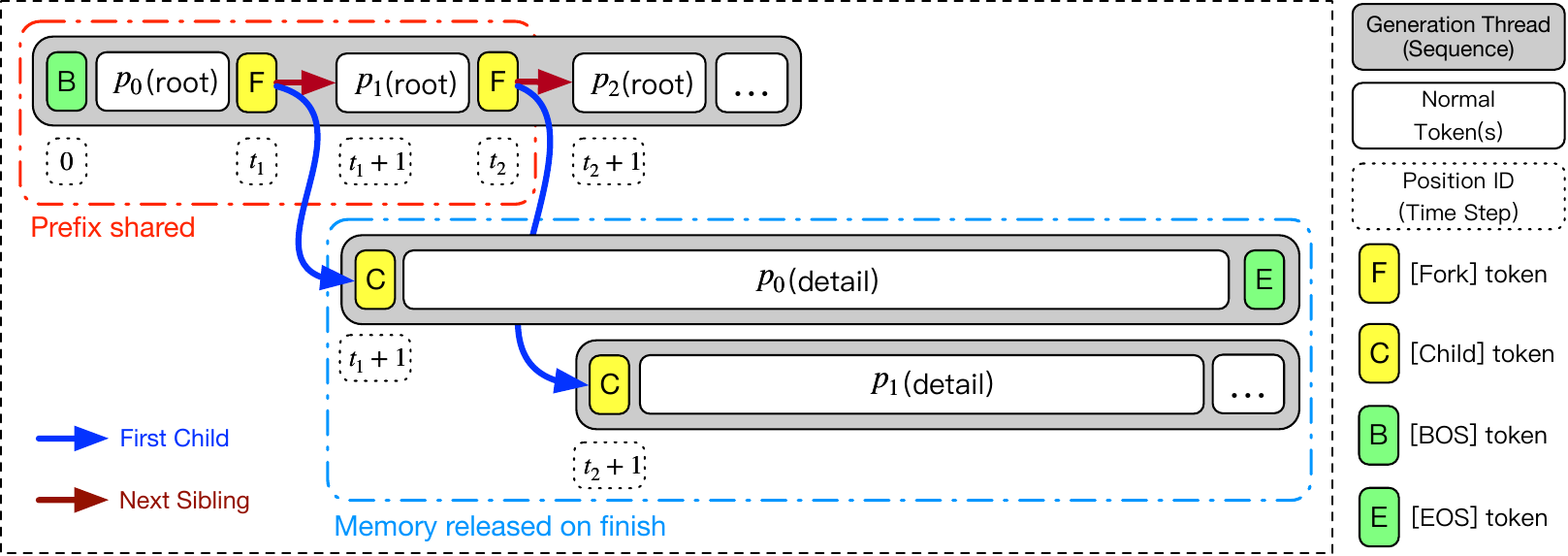}
\end{center}
\caption{\textbf{APAR Decoding.} Suppose we are generating a sequence $p_0-p_1-p_2$. The decoding of $p_1$(root) and $p_0$(detail) are issued in parallel at ($t_1+1$). Similarly, $p_2$(root) and $p_1$(detail) are issued at ($t_2+1$). Prefix tokens are shared and forked generation threads can be released once finished.}
\label{fig:decoding}
\end{figure*}

The key idea of APAR is to make LLMs explicitly aware of such parallelizable structures, and spawn auto-parallel auto-regressive decoding threads accordingly.
Specifically, APAR involves two key components. 
First, we post-train the LLMs with hierarchical text structures, which we referred to as \emph{paragraph nodes} (Section~\ref{method:input_format}). 
Second, we design the decoding algorithm to support parallel decoding operations, including maintaining the hierarchical structure in generation and restoring it into a linear sequence (Section~\ref{method:inference_algo}).

\subsection{Input Format}
\label{method:input_format}

This section introduce the corpora used to fine-tune APAR models, including the tree structure, attention mechanism and control tokens. See Section~\ref{experiments:setup} for data pre-processing details.

\paragraph{Paragraph tree.} As illustrated in Fig~\ref{fig:paragraph_tree}, a paragraph tree is used to represent a sequence with hierarchical structure. Each node in the tree, which is referred to as a \emph{paragraph node} in the sections to follow, denotes a component of the generation response. Each paragraph node has 0 or 2 pointers. One pointer of the paragraph node is referred to as \emph{first child} (\textcolor{blue}{blue} arrow in Fig~\ref{fig:paragraph_tree}), which points to the detailed texts (sub-paragraph) of the paragraph; the other child pointer is \emph{next sibling} (\textcolor{darkred}{red} arrow in Fig~\ref{fig:paragraph_tree}), which points to the next paragraph of the same hierarchical level.

\paragraph{Control tokens.}
To enable the language model to spawn a parallel decoding thread, 2 control tokens are added to the vocabulary.
\begin{itemize}[leftmargin=*,itemsep=0pt,parsep=0.2em,topsep=0.0em,partopsep=0.0em]
    \item \textbf{Fork Identifier.} \fork is used to indicate a parallel-able structure in the response. Models are trained to output a \fork token when they discover that what follows is considered detailed information (or a sub-paragraph) and thus can be decoded together with the next paragraph of the same level. When the inference system detects a \fork token output by the model, it creates a paralleled decoding thread sharing the same prefix until that \fork token. This token works just like the \verb|fork()| system call used in operating systems.
    \item \textbf{Child Identifier.} \child always follows a \fork, and is used to indicate that the following content is the beginning of a sub-paragraph lead by the previous content, like the zero return value of \verb|fork()| in some operating systems. In particular, \child is attended to but is not taken loss in the training process. Thus, the model never learns to output this token, but learns to output the content of the sub-paragraph when \child is injected into the context (i.e. a \fork \child sequence appears in the context). On the other hand, when \fork appears without followed by a \child, the model will generate the next paragraph of the same level.
\end{itemize}

\paragraph{Training attention.} In order for the paragraphs to be generated in a parallel, all nodes only attend to their ancestors, and to themselves with a causal mask, as shown in Fig~\ref{fig:paragraph_tree}. 

\subsection{Decoding Procedures}
\label{method:inference_algo}

\begin{algorithm}[htbp]
\caption{
APAR decoding algorithm. \textproc{IsFinished}($G$) returns True if all sequences in the sequence group $G$ have finished. \textproc{Sample}($\Theta$, $s$) samples the next token for sequence $s$ given language model $\Theta$. A paragraph node $n$ associated with sequence $s$ points to slice $s[\text{start:end}]$ ($s[\text{start:}]$ if end is \emph{null}). \textproc{PNode}(s, $x$) creates a new paragraph node associated with sequence $s$, initializing start=$x$, end=\emph{null}. The current\_node attribute of a sequence is used to record the leaf paragraph node pointing to the end of the sequence, and is maintained (line~\ref{lst:line:being_of_current_node_maintain}$\sim$\ref{lst:line:end_of_current_node_maintain}) every time a new sequence is forked. \textproc{Restore}($r$) recursively traverse the paragraph tree defined by root $r$ in 
root - first\_child - next\_sibling order.
}
\label{alg:apar_decode}
\begin{algorithmic}[1]

\State {\bf Input:} A user prompt sequence $p$, language model $\Theta$.
\State {\bf Output:} Decoded generation sequence $g$.
\State
\State $r \gets$ \Call{PNode}{$p$, $p$.len} 
\State $G \gets \{p\}$  
\State $p$.current\_node $\gets r$ 

\State
\While{not \Call{IsFinished}{G}}
    \State $G\gets$ \Call{AparDecode}{G, $\Theta$}
\EndWhile
\State

\State $g \gets $ \Call{Restore}{$r$} 
\State {\bf return} g

\State

\Function{AparDecode}{$G$, $\Theta$} 
    \For {each sequence $s$ in $G$}
        \If{ $s$.finished = True}
            \State {\bf continue}
        \EndIf
        \State $x \gets $\Call{Sample}{$\Theta$, s}
        \If {$s.\text{last\_token}$ = \fork} 
            \State{$s' \gets s$} 
            \State{$s'.$append\_token(\child)} 
            \State $G\gets G \cup \{s'\}$

            \State $n \gets$ \Call{PNode}{$s$, $s$.len} 
            \State $n' \gets$ \Call{PNode}{$s'$, $s'$.len}
            
            \State $s$.current\_node.end $\gets s$.len  
            \label{lst:line:being_of_current_node_maintain}
            \State $s$.current\_node.next\_sibling $\gets n$ 
            \State $s$.current\_node.first\_child $\gets n'$
            \State $s$.current\_node $\gets n$ 
            \State $s'$.current\_node $\gets n'$ \label{lst:line:end_of_current_node_maintain}
        \EndIf
        \State{$s.$append\_token($x$)} 
        \If{ $x$ = \texttt{[EOS]}  }
            \State $s.$finish $\gets$ True
            \State \Call{ReleaseCache}{s} 
        \EndIf
    \EndFor
    \State {\bf return} $G$
\EndFunction

\end{algorithmic}
\end{algorithm}

An overview of the generation process is illustrated in Fig.~\ref{fig:decoding} and the algorithm is formulated in Algorithm~\ref{alg:apar_decode}. We first introduce the concept of sequence and sequence groups following the implementation in \citet{vllm}, then expound the generating procedures of APAR decoding algorithm.

\paragraph{Sequence and sequence group.} A sequence is defined as an ordered list of tokens. A sequence group is the set of all sequences generated for the same prompt sequence and is initialized with only the prompt sequence. 
In APAR decoding algorithm, each sequence group corresponds to a paragraph tree. 
 
Each sequence, on the other hand, is a generation thread and is associated with a leaf node in the paragraph tree.

\paragraph{Decoding.} As described in Algorithm~\ref{alg:apar_decode}, we start the decoding with the user prompt sequence $p$ and construct a sequence group $G$ initialized as \{$p$\}. We initialize the paragraph tree corresponding to $G$ with root node $r$ and associate sequence $p$ with $r$ (which is now a leaf node). Then, we iterative perform \textproc{AparDecode} on sequence group $G$ until all sequences in $G$ have finished. In the end, the paragraph tree is traversed to restore the sequential output $g$.

Next, we delve into the details for \textproc{AparDecode}, which performs a single decode step for sequence group $G$ with language model $\Theta$. For each unfinished sequence $s$ in $G$,  If the last token is \fork, it means that the model calls for a new generation thread, and now it's time to fork the sequence $s'$. 
The forked sequence shares the same prefix tokens as the parent sequence. When implementation with paged attention~\citep{vllm}, the fork operation creates a shared memory mapping for the shared prefix (the dotted red box in Fig~\ref{fig:decoding}), which \emph{copies at most 1} KV cache block and \emph{shares} all other blocks. 
After the fork operation, $s'$ is appended a forced \child token to identify this sequence to the language model as a child sequence. Also, two new leaf nodes are created and $s$ and $s'$ are set to track the latest leaf nodes. 

Finally, sampled new token $x$ is appended to $s$. If \texttt{[EOS]} token is sampled, the generation of sequence $s$ is considered finished, and the KV cache belonging \emph{only} to $s$ is released (the dotted red box in Fig~\ref{fig:decoding}). 

\subsection{Features} 
\label{method:features}
Based on the aforementioned decoding algorithm and the distribution of user queries, we identify three key features of APAR decoding that give rise to its superior performance in terms of inference latency, throughput, and memory consumption.
    \paragraph{1. Paralleled decoding structure reduces latency.} Through training on paragraph trees, the language model becomes an automatic online miner for parallel-able structure and concurrent generation threads are issued accordingly. The paralleled generation reduces generation steps. In memory-bound scenarios, the latency in each step remains roughly unchanged wrt. different level of decoding parallelism (i.e. dynamic batch size) and the latency can therefore be reduced proportionately (Fig~\ref{fig:generation-speed-memory-bound}).

    \paragraph{2. Early release of child KV cache reduces memory consumption.} In the auto-regressive generation process, the KV cache of all tokens must be retained before the sequence is completely generated. In APAR, however, once a forked sequence (i.e. a generation thread) completes generation, the KV cache belonging only to the forked sequence can be released immediately, while the remaining part of the generation continues. Under the effect of early release strategy, as shown in later Fig~\ref{fig:throughput-memory}, up to $50\%$ of the generation cache can be saved while throughput remains the same.
    
    \paragraph{3. Reduced attention length saves computation.} Auto-regressive generation requires a token to attend to all previously generated tokens. In APAR, on the other hand, a new token only attends to tokens along its path to the root of the paragraph tree, which reduces attention computation in generation. In a heavily-batched generation setting, latency incurred by memory access is amortized by the intense batched computation, make the generation process primarily computation-bound. Thus, the computation reduction in each token results in an improvement in throughput across different memory usages (Fig~\ref{fig:throughput-memory}), as well as reduction in latency with different extents of concurrency (Fig~\ref{fig:latency-concurrency}).

\section{Experiments}

\subsection{Data Pre-processing}
\label{experiments:data}

We adopt one open-sourced version of ShareGPT dataset\footnote{\scriptsize\url{https://huggingface.co/datasets/anon8231489123/ShareGPT_Vicuna_unfiltered}} as instruction corpora. Fine-tuning data are composed as follows.

\paragraph{Ordered list.} Many responses are represented as ordered lists with the pattern of \verb|root - detail| for each bullet point, where \verb|root| is usually an introduction phrase and \verb|detail| is the detailed content of that specific point. Thus, the \verb|root| is extracted as the content for the root node and \verb|detail| as the content in the detail node, as illustrated in Fig~\ref{fig:paragraph_tree}.

\paragraph{Paragraph.} Most of LLM's response paragraphs are structured in a root-and-details format, even when not presented as an ordered list, where the first sentence of the paragraph typically summarizes the main idea of that paragraph. Therefore, we extract the first sentence of the paragraph as the  \verb|root| for that section, while the rest of the content serves as the \verb|detail|. 

\paragraph{Unstructured data.} To accurately extract the hierarchical structure, responses with confusing formats, like code and math data, are excluded in the aforementioned structure extraction process. However, while learning to generate paralleled decoding threads, a model must also learn not to generate \fork in cases where coherent attention is necessary to accurately predict the next token. Therefore, some filtered conversations are added as negative examples to prevent the model from excessive \fork issuing. This portion of data is organized as a single paragraph node with no descendants. 

See Appendix~\ref{appdendix:rules} for detailed procedures and rules used in data pre-processing. 

\begin{figure*}[t]
\begin{center}
\begin{subfigure}{.48\textwidth}
\centering
\includegraphics[width=\textwidth]{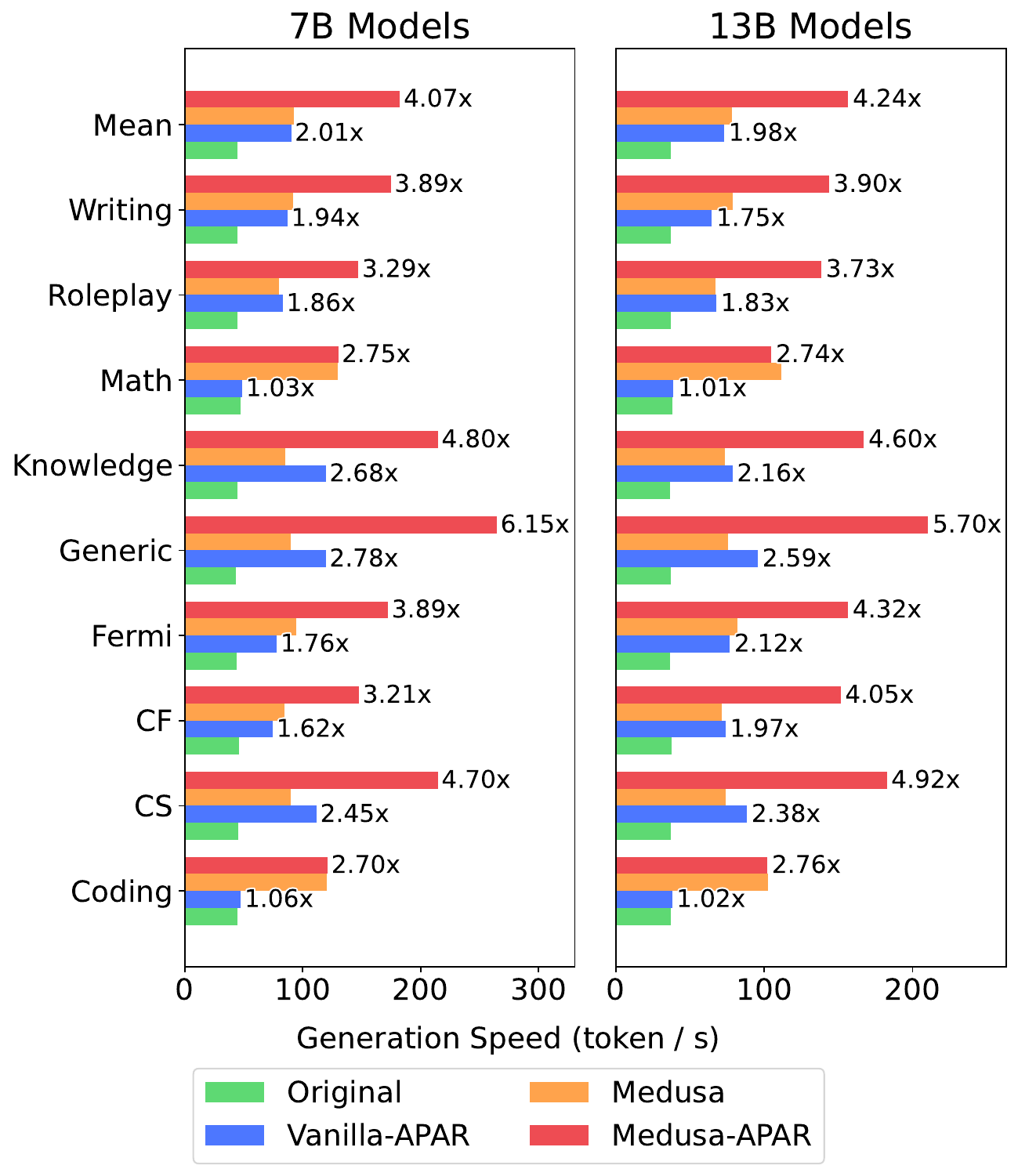}
\caption{Results in Vicuna Bench}
\label{fig:generation-speed-vicuna-bench}
\end{subfigure}
\hfill 
\begin{subfigure}{.48\textwidth}
\centering
\includegraphics[width=\textwidth]{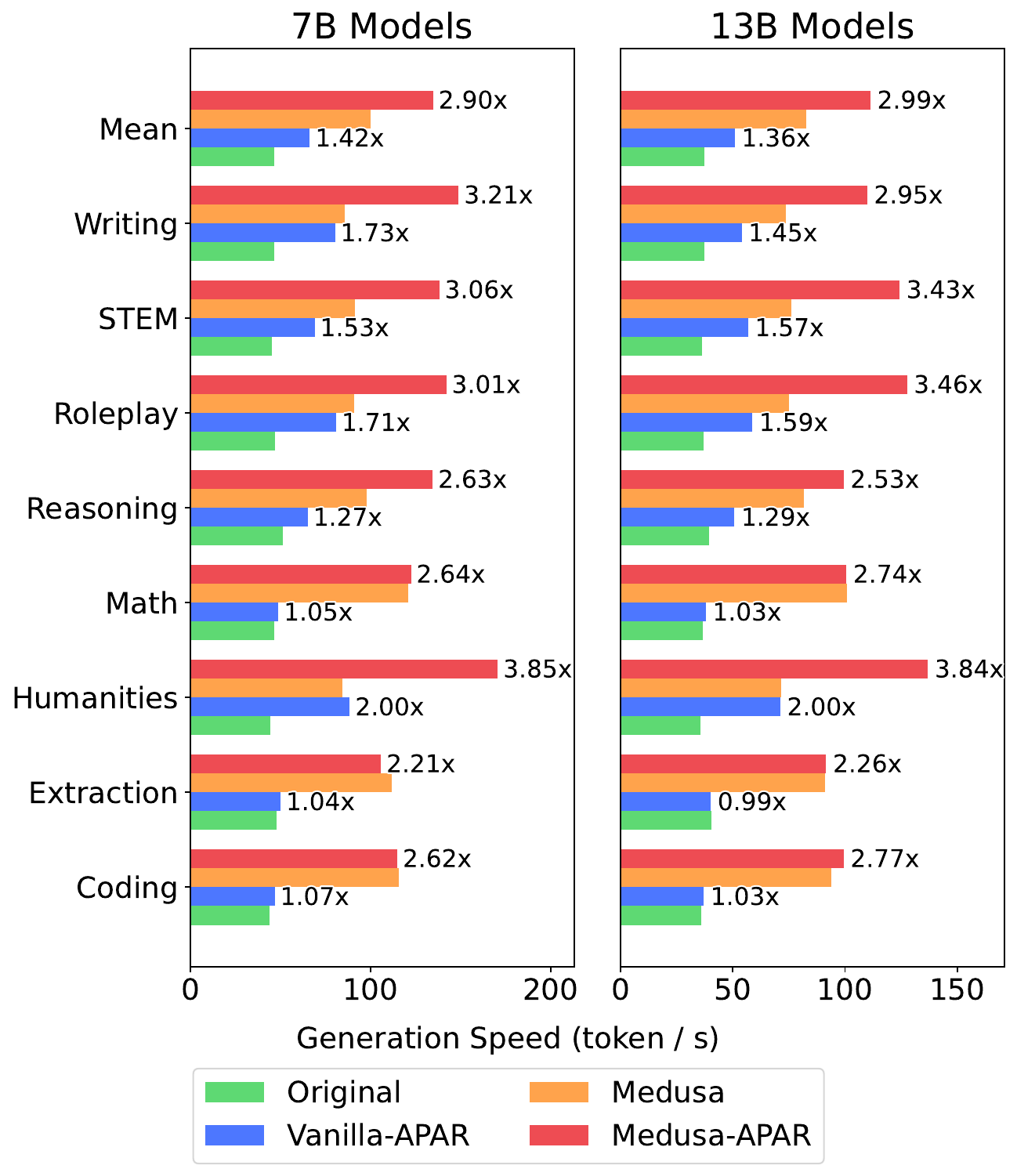}
\caption{Results in MT Bench}
\label{fig:generation-speed-mt-bench}
\end{subfigure}
\end{center}
\caption{\textbf{Generation speed in memory-bound scenario.} Models served on one H800-SXM5-80G GPU with batch size 1.}
\label{fig:generation-speed-memory-bound}
\end{figure*}

\subsection{Experimental Setup}

\label{experiments:setup}

\paragraph{Models.} To evaluate the generation speed, throughtput and qualities, we apply APAR fine-tuning on vicuna-v1.3-\{7B,13B\} models, producing APAR-\{7B,13B\}. In this section, original Vicuna models will be referred to as Original-\{7B,13B\} (O-\{7B,13B\} as abbreviations) and the fine-tuned APAR models will be referred to as APAR-\{7B,13B\} (A-\{7B,13B\} as abbreviations).

\paragraph{Implementation.} We implement 3 settings for evaluation, including 
\begin{itemize}[leftmargin=*,itemsep=0pt,parsep=0.2em,topsep=0.0em,partopsep=0.0em]
    \item \textbf{\vapar}.  \vapar is implemented directly with \verb|transformers|~\citep{wolf-etal-2020-transformers}, which is a widely adopted python deep learning platform for transformer-based models. 
    \item \textbf{\mapar}. \mapar is implemented with Medusa~\citep{medusa}, which is an open-source speculative decoding algorithm that follows the predict - verify paradigm for decoding. Medusa adopts a light-weighed extra language modeling head to predict the next few tokens and verify generation using tree attention. This setting is used to test the combined effect of APAR and speculative decoding algorithm.
    \item \textbf{\bapar}. \bapar is implemented with vLLM~\citep{vllm}, a high-throughput and memory-efficient inference engine using paged-attention mechanism. This setting is used to test APAR on realistic serving situations, where we not only care about the latency but also throughput and memory efficiency.
\end{itemize} 

\paragraph{Training setup.} During the fine-tuning process, we sample from structured (ordered list and paragraph mentioned above, 16k samples) and unstructured data (9k samples) with sampling ratio 1:1. The models are fine-tuned with batch size 128, learning rate $2e\text{-}5$ for 2000 steps. After fine-tuning, we train 2 medusa heads with learning rate 1$e$-3 for 2000 steps using the same data as fine-tuning. See Appendix~\ref{appdendix:hyperparameters} for detailed hyper-parameter settings.

\paragraph{Evaluation datasets.} Several datasets are used to evaluate the generation statistics and quality.
\begin{itemize}
    \item \textbf{Vicuna Bench}~\citep{vicuna2023} is a benchmark for evaluating LLMs on language understanding, reasoning and context awareness. It covers 9 categories and contains 80 single-turn queries. For a clearer layout, we abbreviate 2 long category names in Vicuna Bench in the following figures and tables, i.e. Commonsense to CS, Counterfactual to CF.
    
    \item \textbf{MT Bench}~\citep{mt-bench} is a benchmark consisting of 80 multi-turn questions. It covers 8 common categories and can be used to evaluate the multi-turn conversation and instruction-following ability of LLMs.

    \item \textbf{APAR Test Set} consists of 1000 user queries sampled from ShareGPT dataset to simulate the query distribution in realistic deployment scene using the same rule we extract structured training data.
    Because of its large quantity, it's would be too expensive to evaluate generation quality for all test set queries on all models. Thus, APAR test set is only used for measurement of generation statistics. 
\end{itemize}

\subsection{Results in Memory-Bound Scenarios}
\label{experiments:speed-mem}

We inspect how APAR reduces generation latency in a memory-bound (i.e. small batch size) scenario, as well as its combined acceleration effect with speculative decoding. Considering that the the model is re-trained and the output length can be different on the same prompt, we normalize generation latency with generated tokens, adopting tokens per second as the metric for generation speed. The results are reported with batch size fixed as 1 and prefix sharing is not enabled.

As shown in Fig~\ref{fig:generation-speed-memory-bound}, \vapar  achieves $2\times$ average speed up in Vicuna Bench and $1.4\times$ average speed up on MT Bench. APAR models learns to spawn parallel generation thread in and only in categories that exists a parallel-able structure. For instance, APAR-\{7B,13B\} seldom try to issue parallel a generation threads in coding and math related queries, which typically requires careful step by step reasoning or rigorous formats, resulting in no speed up. On the other hand, on categories like common-sense, generic and knowledge, the speed up is significant. When combined with speculative decoding, \mapar achieves an impressive $4 \times$ average speed up in Vicuna Bench and $2.9 \times $ average speed up in MT Bench, demonstrating strong reduction in generation latency.

\subsection{Results in High-Throughput Scenarios}
\label{experiments:speed-th}

\begin{figure}[h]

\begin{center}
\begin{subfigure}{.45\textwidth}
\centering
\includegraphics[width=\textwidth]{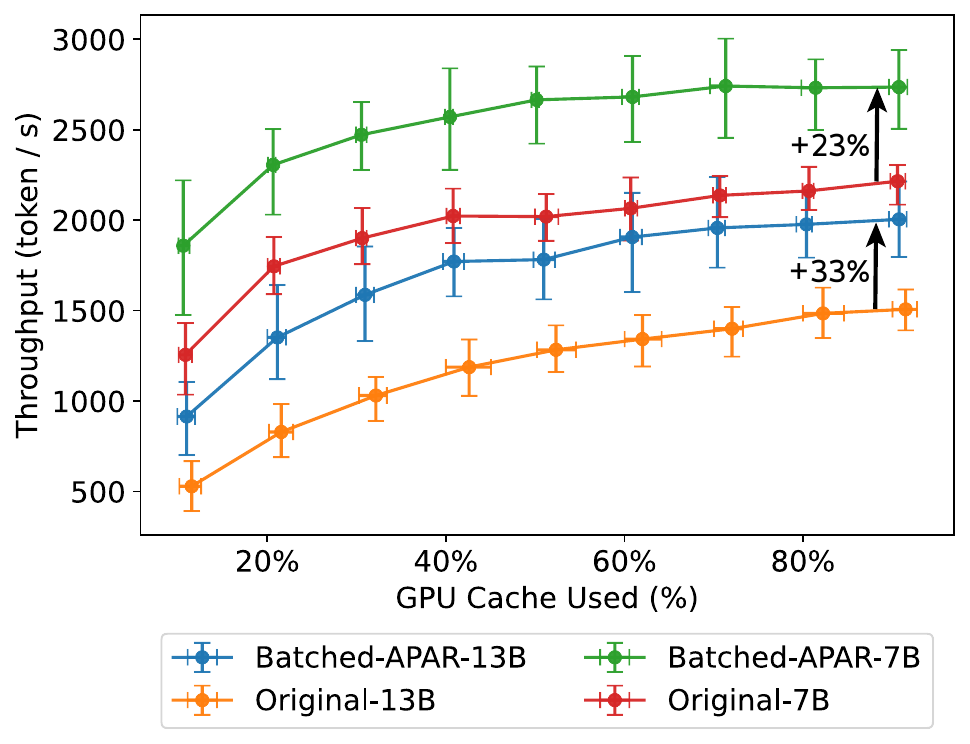}
\caption{Throughput wrt. cache memory usage}
\label{fig:throughput-memory}
\end{subfigure}

\begin{subfigure}{.45\textwidth}
\centering
\includegraphics[width=\textwidth]{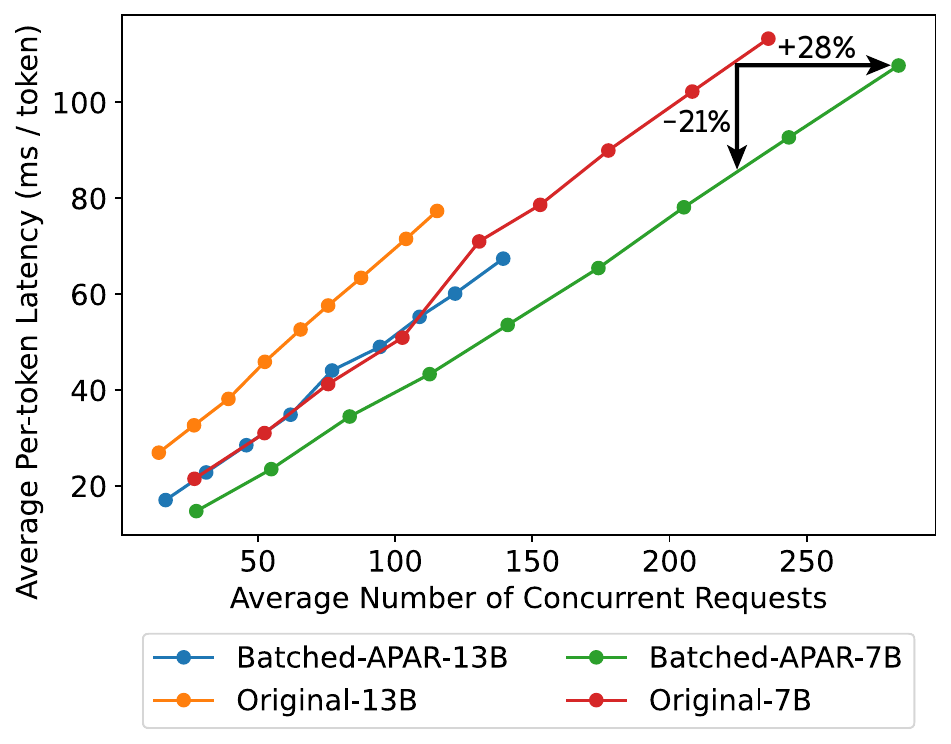}
\caption{End-to-end decode latency wrt. concurrency}
\label{fig:latency-concurrency}
\end{subfigure}

\end{center}
\caption{\textbf{Serving statistics of \bapar.} Models served on one A100-SXM4-80G GPU. Error bars in 
(b)
omitted for a clearer view. }
\label{fig:generation-computation-bound}
\end{figure}

In high performance serving situations, increasing throughput and reducing serving memory are also important. We use \bapar to serve the queries in APAR test set with different amount of GPU memory available. An overview of generation throughput and per-token latency is summarized in Fig~\ref{fig:generation-computation-bound}. The dots in the plot show the mean value and error bars represent the 25\% and 75\% percentile in each setting. 

As shown in Fig~\ref{fig:throughput-memory}, the throughput of \bapar models surpass the maximum throughput of original models with only 20\% of the KV Cache used, demonstrating memory efficiency. When using similar amount of memory, throughput is consistently increase by 20\%$\sim$ 70\% across different cache usages.
\bapar models also demonstrate remarkable latency reduction in computation bound scenarios. As shown in Fig~\ref{fig:latency-concurrency}, \bapar reduces 20\%$\sim$35\% average latency when serving the same amount of concurrent requests. The latency of \bapar-13B is even similar to the Original-7B model.

The improvement in latency and throughput can be best explained by feature 2 and 3 as described in Section~\ref{method:features}. We quantitatively measure how much computation and cache memory can be saved by using APAR decoding algorithm. We adopt the following metrics.

\begin{itemize}
    \item \textbf{Max cached tokens} is defined as the maximum number of KV cache slots required for generating a response. For auto-regressive generation, prompt tokens and all generated tokens need to be cached before generating \texttt{[EOS]} token.

    \item \textbf{Attended tokens} is defined as the number of tokens attended to when predicting a specific token. For auto-regressive generation, all preceding tokens is needed when predicting a next token.
\end{itemize}

\begin{table}[t]
\centering
\caption{Average Max Cached Tokens}
\label{tab:mean-saved-token-cache}
\renewcommand\tabcolsep{5pt}
\begin{tabular}{cc|ccc}
\toprule
Benchmark                                     & \multicolumn{1}{c|}{Model} & \multicolumn{1}{c}{APAR} & \multicolumn{1}{c}{Flatten} & \multicolumn{1}{c}{Saved} \\ \midrule
\multirow{2}{*}{Vicuna}                 & A-7B            & 303.2                    & 417.2                  & 27.3\%                    \\
                                              & A-13B           & 306.7                    & 419.3                  & 26.8\%                    \\ \midrule
\multicolumn{1}{c}{\multirow{2}{*}{MT}} & A-7B            & 502.1                    & 571.1                  & 12.1\%                    \\
\multicolumn{1}{c}{}                          & A-13B           & 513.3                    & 590.6                  & 13.1\%                    \\ \bottomrule
\end{tabular}
\end{table}
\begin{table}[t]
\centering
\renewcommand\tabcolsep{5pt}
\caption{Average Attended Tokens}
\label{tab:mean-saved-attened-tokens}
\begin{tabular}{cc|ccc}
\toprule
Benchmark                                     & \multicolumn{1}{c|}{Model} & \multicolumn{1}{c}{APAR} & \multicolumn{1}{c}{Flatten} & \multicolumn{1}{c}{Saved} \\ \midrule
\multirow{2}{*}{Vicuna} & A-7B  & 166.2  & 256.4   & 35.2\%   \\
                        & A-13B & 166.6  & 255.9   & 34.9\%   \\ \midrule
\multicolumn{1}{c}{\multirow{2}{*}{MT}} & A-7B  & 417.5 & 496.5 & 15.9\%  \\
\multicolumn{1}{c}{}                    & A-13B & 428.0 & 514.2 & 16.8\%  \\ \bottomrule
\end{tabular}
\end{table}

Since the response length is difference between APAR models and original models, we flatten the paragraph tree generated by APAR models as the reference output. When calculating average, we exclude categories that are not accelerated by APAR, i.e. Coding, Extraction and Math. 

As summarized in Table~\ref{tab:mean-saved-token-cache} and Table~\ref{tab:mean-saved-attened-tokens}, compared with flattened results, APAR reduces max cached tokens by 12\%$\sim$27\% and reduced attended tokens by 15\%$\sim$35\%. Detailed results for all categories are reported in Appendix~\ref{appendix:attended-token} and Appendix~\ref{appendix:token-caches}.


\subsection{Generation Quality}
\label{experiments:quailty}


\begin{table}[t]
\centering

\caption{Generation Quality in MT Bench}
\label{tab:mt_bench}
\renewcommand\tabcolsep{5pt}
\begin{tabular}{c|cc|cc}
\toprule
\textbf{Task} & \textbf{O-7B} & \textbf{A-7B} & \textbf{O-13B} & \textbf{A-13B} \\ \midrule
Coding & 2.60 & 2.60 & 3.70 & 3.70 \\
Extraction & 4.90 & 5.75  & 5.35 & 4.85 \\
Humanities & 9.45 & 9.40  & 9.45 & 9.55  \\
Math & 2.70 & 2.00  & 2.60 & 2.65  \\
Writing & 7.90 & 7.15  & 8.40 & 7.88 \\
Roleplay & 6.35 & 7.25  & 7.67 & 7.58  \\
Reasoning & 4.85 & 5.40  & 5.95 & 5.85  \\
Stem & 7.92 & 7.15  & 7.67 & 7.80  \\
\midrule
\textbf{Mean} & 5.83 & 5.92  & 6.35 & 6.24  \\
\bottomrule
\end{tabular}

\end{table}
\begin{table}[t]
\centering
\renewcommand\tabcolsep{5pt}
\caption{Generation Quality in Vicuna Bench}
\label{tab:vicuna_bench}
\begin{tabular}{c|cc|cc}
\toprule
\textbf{Task}          & \textbf{O-7B} & \textbf{A-7B} & \textbf{O-13B} & \textbf{A-13B} \\ 
\midrule
Coding & 3.86 & 3.29  & 6.14 & 3.71  \\
CS & 9.40 & 9.50  & 9.50 & 9.60  \\
CF & 8.05 & 8.00  & 8.65 & 9.00  \\
Fermi & 6.80 & 6.90  & 7.35 & 6.60  \\
Generic & 9.45 & 9.50  & 9.30 & 9.50  \\
Knowledge & 9.40 & 9.30  & 9.60 & 9.35  \\
Math & 1.67 & 2.00  & 1.67 & 4.67  \\
Roleplay & 8.80 & 8.80  & 8.90 & 8.75  \\
Writing & 9.50 & 9.00 & 9.50 & 9.00 \\
\midrule
\textbf{Mean} & 8.08 & 7.99  & 8.45 & 8.29  \\
\bottomrule
\end{tabular}

\end{table}

To measure the generation quality of APAR models compared with  original models, we adopt MT Bench and Vicuna Bench as  evaluation framework. For each response, we provide GPT-4 with conversation history, user query and model responses, asking GPT-4 to grade the response with a score ranging from 1 to 10 and we follow the prompt template used by \citet{mt-bench}.

The quality scores of each category are summarized in Table~\ref{tab:mt_bench} and Table~\ref{tab:vicuna_bench}. Compared with original models, APAR models differs by -2\%$\sim$+2\% in MT Bench and Vicuna Bench overall scores, showing negligible overall quality change.

\section{Related Works}
\label{related-works}

This section discuss the difference and connection of APAR with prior works concerning inference acceleration.

\paragraph{Optimized computation.} Optimizations on operators~\citep{dao2022flashattention} and computational graphs~\citep{deepspeed-inference} are active research fields. Model compression is widely used in deployment, like quantization~\citep{dettmers2022llmint8, frantar-gptq} and pruning~\citep{frantar2023sparsegpt, ma2023llmpruner}. Another line of works modifies the model architecture, including efficient attention~\citep{kitaev2020reformer} for computation complexity and multi-query attention~\citep{shazeer2019fast} for optimized IO. 
Different from prior works, APAR makes no modification to operators or model architecture but reduces computation by adopting attention tree structure. APRA is thus \emph{orthogonal to} and can be applied jointly with the aforementioned works.

\paragraph{Improved parallelism.} 
Scheduling strategies, including dynamic batching~\citep{orca} and paged-attention~\citep{vllm}, improve maximum generation throughput. Another stream of works explores speculative decoding (SD)~\citep{spculative-decoding,infwithref,medusa}, which verifies multiple speculated tokens in parallel, reducing generation latency in small batch sizes. Non-auto-regressive generation~\citep{gu2018nonautoregressive} propose to sample multiple generation tokens in parallel, which typically requires re-training and applies to restricted scenarios. 
APAR can be \textit{conveniently combined} with efficient scheduling and SD methods to achieve augmented efficiency as demonstrated by \mapar and \bapar. Different from previous methods, APAR propose to exploit the \emph{intrinsic organization ability} of LLMs to automatically issue paralleled generation threads, and is applicable to multiple scenarios. Notably, SoT~\citep{sot} proposes to enable parallelism by prompting, which generates the skeleton of the response and then expands each point in parallel. Different from SoT, which entails an external classifier and re-computation of KV cache between stages, APAR requires negligible extra computation (2 control tokens for a thread) and no re-computation, and thus does not compromise generation throughput.

\section{Conclusion}
\label{conclusion}

This paper introduces APAR, a new decoding method that allows LLMs to autonomously structure the decoding process and dynamically create parallel decoding threads, without compromising the generation quality. 
APAR not only enhances parallelism in generation, but also reduces the computation and KV cache memory consumption. 
Experiments show that APAR can be seamlessly integrated with existing inference frameworks, significantly lowering the generation latency across various scenarios while improving serving throughput in situations involving extreme batch sizes and concurrency levels.



\bibliography{acl_latex}

\clearpage

\appendix

\section{Training Hyper-parameters}
\label{appdendix:hyperparameters}

\subsection{APAR Models}

\begin{table}[h]
\centering
\caption{Training Hyper-parameters for APAR Models}
\label{tab:my-table}
\begin{tabular}{ll}
\toprule
Hyper-parameter     & Value  \\ \midrule
batch size         & 128    \\
data type          & bf16   \\
training step      & 2000   \\
learning rate      & 2e-5   \\
weight decay       & 0      \\
warm-up ratio      & 0.03   \\
lr decay schedule  & cosine \\
context length & 2048   \\ \bottomrule
\end{tabular}
\end{table}

\subsection{Medusa Heads}

The following hyper-parameters are used to train the medusa heads for speculative decoding. Only medusa heads are trained and the reset of the language model remains frozen.

\begin{table}[h]
\centering
\caption{Training Hyper-parameters for Medusa Heads}
\label{tab:my-table}
\begin{tabular}{ll}
\toprule
Hyper-parameter     & Value  \\ \midrule
batch size         & 128    \\
data type          & bf16   \\
training step      & 2000   \\
learning rate      & 1e-3   \\
weight decay       & 0      \\
warm-up ratio      & 0.1    \\
lr decay schedule  & cosine \\
context length & 2048   \\ 
\# of Medusa heads & 2   \\ 
\# of layers per Medusa head  & 2   \\ 

\bottomrule
\end{tabular}
\end{table}

\section{Rules for Extracting Structured Data}
\label{appdendix:rules}

The process and rules to determine and extract the structure of each assistant response are outlined as follows.

\begin{enumerate}
    \item Try to extract the response as ordered list.
    \begin{enumerate}
        \item Use regular expression like \\ 
        \verb|(\d+\.)\s+(.+?):(.+?)| \\
        to extract individual numeric points.
        \item If 
        \begin{enumerate}
            \item the regular expression does not does not match at least 3 numeric points, or
            \item any of the content of the numeric points are less then 10 characters
        \end{enumerate}
        the response is not consider as a valid ordered list.
    \end{enumerate}
    \item If the response is not consider as a valid ordered list, try to extract the response as multiple paragraph. 
    \begin{enumerate}
        \item Use two consecutive \verb|\n| to divide the entire response into paragraphs and extract the first sentence of each paragraph. Paragraphs with only one sentence is skipped.
    \end{enumerate}
    \item If ambiguous patterns, including code blocks, math expressions, URLs, etc, exists in the response, the, or if the response fails to match the above 2 criteria, the response is considered unstructured.
\end{enumerate}

Pre-processing code will be made public in the project repository.

\section{Generation Speed Evaluation Setup}
\label{appdendix:speed-setup}

\subsection{\vapar and \mapar}

\vapar and \mapar are implementation directly from \texttt{transformers} package. To keep KV cache contiguous, prefix sharing is not enabled and the prefix KV caches are copied when a new generation thread is forked. The implementation of \mapar is adopt from its official repository. When different number of tokens are accepted across batches, the longest accepted length is adopted and the mis-predicted slots are masked out in attention calculation. The evaluation batch-size is fixed as 1 and the pre-filling time is not measured when calculating generation speed.

\subsection{\bapar}

To measure the maximum throughput of each generation method, GPU cache utilization is set to $0.1,\cdots,0.9$ for each setting and the system is profiled every 3 seconds. For a stable performance, we ignore the terminal samples when no requests are pending and ignore the first $\frac 1 3$ samples as warm-up when calculating mean and percentiles. 

All 1k requests are push into waiting queue in the beginning of the performance test. We limit the max number of concurrent requests to 350 for 7B models and 180 for 13B models. The reason for this limit is that in the beginning, the average sequence length is relatively sort and the system is prone to excessively accept requests, leading to frequent swapping and recompute of KV cache. Note that this concurrency limit mainly takes effect in warm-up stage, which is ignore in the calculation for mean and percentage. The concurrency limit is much larger than the maximum in the average concurrent request as shown in Fig~\ref{fig:latency-concurrency}.

\section{Generation Speed Details}
\label{appendix:speed-details}

\subsection{Parallel Generation Statistics}
\label{appendix:parallel-generation-statistics}

We measure how many parallel threads are issued in generation across benchmarks and categories. Results are reported in Table~\ref{tab:parallel-generation-statistics-vicuna-bench} and Table~\ref{tab:parallel-generation-statistics-mt-bench}. \#T stands for average number of generation threads, \%P stand for ratio of parallel-able response, i.e. response that has at least 2 generation threads.

\begin{table}
\centering
\caption{Parallel Generation Statistics in Vicuna Bench}
\label{tab:parallel-generation-statistics-vicuna-bench}
\begin{tabular}{c|cc|cc}
\toprule
\multirow{2}{*}{Task} & \multicolumn{2}{c|}{APAR-7B} & \multicolumn{2}{c}{APAR-13B} \\
                      & \#T & \%P & \#T & \%P \\
\midrule
Coding & 1.0 & 0.0 & 1.0 & 0.0 \\
CS & 6.0 & 1.0 & 6.5 & 1.0 \\
CF & 4.1 & 0.9 & 5.4 & 1.0 \\
Fermi & 5.3 & 0.9 & 5.5 & 0.9 \\
Generic & 7.7 & 1.0 & 7.7 & 1.0 \\
Knowledge & 6.8 & 1.0 & 5.6 & 1.0 \\
Math & 1.0 & 0.0 & 1.0 & 0.0 \\
Roleplay & 4.5 & 0.8 & 5.1 & 1.0 \\
Writing & 5.4 & 0.9 & 4.6 & 0.7 \\
\midrule
Mean & 5.1 & 0.81 & 5.2 & 0.82 \\
\bottomrule
\end{tabular}
\end{table}

\begin{table}
\centering
\caption{Parallel Generation Statistics in MT Bench}
\label{tab:parallel-generation-statistics-mt-bench}
\begin{tabular}{c|cc|cc}
\toprule
\multirow{2}{*}{Task} & \multicolumn{2}{c|}{APAR-7B} & \multicolumn{2}{c}{APAR-13B} \\
                      & \#T & \%P & \#T & \%P \\
\midrule
Coding & 1.0 & 0.0 & 1.0 & 0.0 \\
Extraction & 1.0 & 0.0 & 1.3 & 0.1 \\
Humanities & 5.0 & 0.7 & 6.1 & 0.8 \\
Math & 1.1 & 0.1 & 1.0 & 0.0 \\
Reasoning & 2.0 & 0.3 & 2.5 & 0.6 \\
Roleplay & 4.1 & 0.8 & 4.0 & 0.7 \\
STEM & 3.4 & 0.6 & 3.6 & 0.6 \\
Writing & 3.7 & 0.6 & 3.4 & 0.6 \\
\midrule
Mean & 2.7 & 0.37 & 2.9 & 0.41 \\
\bottomrule
\end{tabular}
\end{table}

\subsection{Max Cached Tokens}
\label{appendix:token-caches}

Detailed results of max cached tokens across benchmarks and categories are reported in Table~\ref{tab:saved-token-cache-for-vanilla-apar-7b-on-vicuna-bench}, Table~\ref{tab:saved-token-cache-for-vanilla-apar-13b-on-vicuna-bench}, ~\ref{tab:saved-token-cache-for-vanilla-apar-7b-on-mt-bench} and Table~\ref{tab:saved-token-cache-for-vanilla-apar-13b-on-mt-bench}, . Mean value of all categories and categories accelerated by APAR are reported.

\begin{table}
\centering
\caption{Max Cached Tokens (Vanilla-APAR-7B on Vicuna Bench)}
\label{tab:saved-token-cache-for-vanilla-apar-7b-on-vicuna-bench}
\begin{tabular}{c|ccc}
\toprule
Task & APAR & AR & Saved \\
\midrule
Coding & 378.0 & 378.0 &  0.0\% \\
CS & 296.6 & 417.7 &  29.0\% \\
CF & 245.6 & 312.4 &  21.4\% \\
Fermi & 365.3 & 496.8 &  26.5\% \\
Generic & 251.5 & 388.8 &  35.3\% \\
Knowledge & 345.0 & 489.9 &  29.6\% \\
Math & 210.7 & 210.7 &  0.0\% \\
Roleplay & 292.5 & 364.1 &  19.7\% \\
Writing & 325.6 & 450.4 &  27.7\% \\
\midrule
Mean & 306.2 & 406.0 &  24.6\% \\
Mean(>1\%) & 303.2 & 417.2 &  27.3\% \\
\bottomrule
\end{tabular}
\end{table}

\begin{table}
\centering
\caption{Max Cached Tokens (Vanilla-APAR-13B on Vicuna Bench)}
\label{tab:saved-token-cache-for-vanilla-apar-13b-on-vicuna-bench}
\begin{tabular}{c|ccc}
\toprule
Task & APAR & AR & Saved \\
\midrule
Coding & 370.0 & 370.0 &  0.0\% \\
CS & 287.6 & 416.9 &  31.0\% \\
CF & 276.9 & 383.0 &  27.7\% \\
Fermi & 401.0 & 529.7 &  24.3\% \\
Generic & 260.7 & 389.2 &  33.0\% \\
Knowledge & 313.2 & 419.7 &  25.4\% \\
Math & 203.3 & 203.3 &  0.0\% \\
Roleplay & 254.3 & 346.5 &  26.6\% \\
Writing & 353.5 & 449.8 &  21.4\% \\
\midrule
Mean & 308.4 & 406.9 &  24.2\% \\
Mean(>1\%) & 306.7 & 419.3 &  26.8\% \\
\bottomrule
\end{tabular}
\end{table}

\begin{table}
\centering

\caption{Max Cached Tokens (Vanilla-APAR-7B on MT Bench)}
\label{tab:saved-token-cache-for-vanilla-apar-7b-on-mt-bench}
\begin{tabular}{c|ccc}
\toprule
Task & APAR & AR & Saved \\
\midrule
Coding & 744.8 & 744.8 &  0.0\% \\
Extraction & 567.2 & 567.2 &  0.0\% \\
Humanities & 647.0 & 757.3 &  14.6\% \\
Math & 408.5 & 410.4 &  0.4\% \\
Reasoning & 314.9 & 335.4 &  6.1\% \\
Roleplay & 461.2 & 533.6 &  13.6\% \\
STEM & 574.2 & 641.7 &  10.5\% \\
Writing & 513.0 & 587.6 &  12.7\% \\
\midrule
Mean & 529.6 & 573.3 &  7.6\% \\
Mean(>1\%) & 502.1 & 571.1 &  12.1\% \\
\bottomrule
\end{tabular}
\end{table}

\begin{table}
\centering

\caption{Max Cached Tokens (Vanilla-APAR-13B on MT Bench)}
\label{tab:saved-token-cache-for-vanilla-apar-13b-on-mt-bench}
\begin{tabular}{c|ccc}
\toprule
Task & APAR & AR & Saved \\
\midrule
Coding & 589.6 & 589.6 &  0.0\% \\
Extraction & 640.3 & 641.7 &  0.2\% \\
Humanities & 676.1 & 822.5 &  17.8\% \\
Math & 451.9 & 451.9 &  0.0\% \\
Reasoning & 318.2 & 344.8 &  7.7\% \\
Roleplay & 484.6 & 548.5 &  11.7\% \\
STEM & 591.1 & 677.5 &  12.8\% \\
Writing & 496.4 & 559.6 &  11.3\% \\
\midrule
Mean & 530.3 & 579.1 &  8.4\% \\
Mean(>1\%) & 513.3 & 590.6 &  13.1\% \\
\bottomrule
\end{tabular}
\end{table}

\subsection{Attended Tokens}
\label{appendix:attended-token}

Detailed results of attend tokens across benchmarks and categories are reported in Table~\ref{tab:saved-attended-tokens-for-vanilla-apar-7b-on-vicuna-bench}, Table~\ref{tab:saved-attended-tokens-for-vanilla-apar-13b-on-vicuna-bench}, Table~\ref{tab:saved-attended-tokens-for-vanilla-apar-7b-on-mt-bench} and Table~\ref{tab:saved-attended-tokens-for-vanilla-apar-13b-on-mt-bench}. Mean value of all categories and categories accelerated by APAR are reported.

\begin{table}
\centering

\caption{Attended Tokens (Vanilla-APAR-7B on Vicuna Bench)}
\label{tab:saved-attended-tokens-for-vanilla-apar-7b-on-vicuna-bench}
\begin{tabular}{c|ccc}
\toprule
Task & APAR & AR & Saved \\
\midrule
Coding & 221.6 & 221.6 &  0.0\% \\
CS & 152.0 & 244.6 &  37.9\% \\
CF & 142.0 & 185.3 &  23.4\% \\
Fermi & 214.6 & 309.6 &  30.7\% \\
Generic & 126.3 & 222.2 &  43.2\% \\
Knowledge & 154.7 & 290.3 &  46.7\% \\
Math & 139.7 & 139.7 &  0.0\% \\
Roleplay & 166.8 & 229.0 &  27.2\% \\
Writing & 189.1 & 270.0 &  30.0\% \\
\midrule
Mean & 170.3 & 251.7 &  32.4\% \\
Mean(>1\%) & 166.2 & 256.4 &  35.2\% \\
\bottomrule
\end{tabular}
\end{table}

\begin{table}
\centering

\caption{Attended Tokens (Vanilla-APAR-13B on Vicuna Bench)}
\label{tab:saved-attended-tokens-for-vanilla-apar-13b-on-vicuna-bench}
\begin{tabular}{c|ccc}
\toprule
Task & APAR & AR & Saved \\
\midrule
Coding & 222.0 & 222.0 &  0.0\% \\
CS & 151.3 & 248.7 &  39.2\% \\
CF & 144.6 & 224.2 &  35.5\% \\
Fermi & 204.9 & 325.7 &  37.1\% \\
Generic & 125.3 & 221.8 &  43.5\% \\
Knowledge & 158.6 & 248.8 &  36.3\% \\
Math & 137.3 & 137.3 &  0.0\% \\
Roleplay & 150.3 & 217.1 &  30.8\% \\
Writing & 209.0 & 271.8 &  23.1\% \\
\midrule
Mean & 170.5 & 251.4 &  32.2\% \\
Mean(>1\%) & 166.6 & 255.9 &  34.9\% \\
\bottomrule
\end{tabular}
\end{table}

\begin{table}
\centering

\caption{Attended Tokens (Vanilla-APAR-7B on MT Bench)}
\label{tab:saved-attended-tokens-for-vanilla-apar-7b-on-mt-bench}
\begin{tabular}{c|ccc}
\toprule
Task & APAR & AR & Saved \\
\midrule
Coding & 577.4 & 577.4 &  0.0\% \\
Extraction & 570.8 & 570.8 &  0.0\% \\
Humanities & 456.5 & 556.0 &  17.9\% \\
Math & 616.0 & 616.9 &  0.1\% \\
Reasoning & 286.7 & 322.0 &  11.0\% \\
Roleplay & 404.3 & 477.7 &  15.4\% \\
STEM & 413.0 & 473.8 &  12.8\% \\
Writing & 437.9 & 533.1 &  17.8\% \\
\midrule
Mean & 477.0 & 528.3 &  9.7\% \\
Mean(>1\%) & 417.5 & 496.5 &  15.9\% \\
\bottomrule
\end{tabular}
\end{table}

\begin{table}
\centering

\caption{Attended Tokens (Vanilla-APAR-13B on MT Bench)}
\label{tab:saved-attended-tokens-for-vanilla-apar-13b-on-mt-bench}
\begin{tabular}{c|ccc}
\toprule
Task & APAR & AR & Saved \\
\midrule
Coding & 449.9 & 449.9 &  0.0\% \\
Extraction & 767.8 & 769.2 &  0.2\% \\
Humanities & 507.0 & 636.0 &  20.3\% \\
Math & 601.3 & 601.3 &  0.0\% \\
Reasoning & 314.3 & 350.1 &  10.2\% \\
Roleplay & 404.8 & 464.9 &  12.9\% \\
STEM & 445.3 & 521.1 &  14.6\% \\
Writing & 360.8 & 442.7 &  18.5\% \\
\midrule
Mean & 482.3 & 539.6 &  10.6\% \\
Mean(>1\%) & 428.0 & 514.2 &  16.8\% \\
\bottomrule
\end{tabular}
\end{table}

\subsection{Response Length}

\begin{figure}
    \centering
    \includegraphics[width=.48\textwidth]{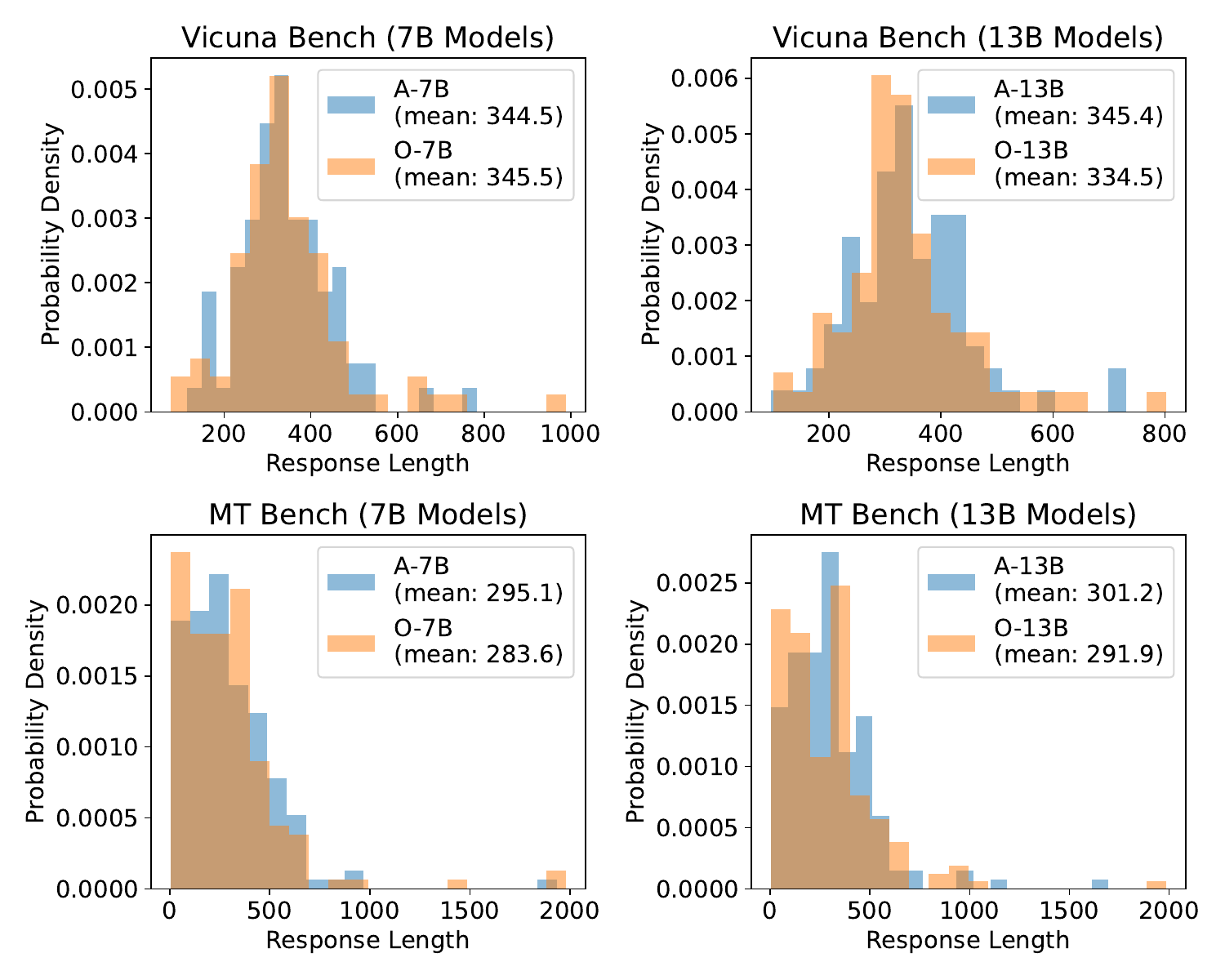}
    \caption{Response Length on Vicuna Bench and MT Bench}
    \label{fig:response-length}
\end{figure}

Apart from generation quality, we also analyze the response length distribution of model before and after fine-tuning in Fig~\ref{fig:response-length}. The average length varies from -0.3\%$\sim$+4.0\% compared with respective original model and the distributions highly overlap. This indicate that if fine-tuned with the same material, APAR does not significantly affect the generation length distribution.


\end{document}